\documentclass[conference]{IEEEtran}
\usepackage{algorithmic}
\usepackage{array}
\usepackage{verbatim}
\usepackage{amsthm}
\usepackage{amsmath,amssymb}
\usepackage{graphicx}
\usepackage{framed}
\usepackage[tight,footnotesize]{subfigure}
\usepackage{listings}
\usepackage{tikz}
\usepackage{tabularx}

\usepackage{color}
\usepackage{url}
\usepackage{algorithm2e}

\definecolor{olivegreen}{rgb}{0.2,0.8,0.5}
\definecolor{grey}{rgb}{0.5,0.5,0.5}

\lstdefinelanguage{ttl}{
	sensitive=true,
	showspaces=false,
	showstringspaces=false,
	keywords={author, birthPlace, birthDate, broader,subject, label, completionDate, type, genre, philosophicalSchool,publicationDate, mainInterest, movement},
	morecomment=[l][\color{grey}]{--},
	morestring=[b][\color{blue}]\",
	morecomment=[s][\color{olivegreen}]{<}{>},
	alsoletter={-,<,>},
	emph={dbo,dbr,dc, Category, rdf, rdfs, skos}, emphstyle=\itshape,
	emph={[2]fun,cat, lincat,lin},emphstyle={[2]\color{red}}
}

\ifCLASSOPTIONcompsoc
\usepackage[caption=false,font=normalsize,labelfont=sf,textfont=sf]{subfig}
\else
\usepackage[caption=false,font=footnotesize]{subfig}
\fi

\newtheorem{example}{Example}

\begin{document}

\IEEEoverridecommandlockouts\IEEEpubid{\makebox[\columnwidth]{978-1-4673-8200-7/15 
\$31.00~\copyright{} 2015 IEEE \hfill} \hspace{\columnsep}\makebox[\columnwidth]{ }}

\title{Information retrieval in folktales using natural language processing}
	
\author{\IEEEauthorblockN{Adrian Groza and Lidia Corde\IEEEauthorrefmark{1}
\IEEEauthorblockA{\IEEEauthorrefmark{1}Intelligent Systems Group,\\Department of Computer Science,\\ Technical University of Cluj-Napoca, Romania\\
	Adrian.Groza@cs.utcluj.ro,Lidia.Corde@cs-gw.utcluj.ro\\ 
	}
	}
	}
	\maketitle
	
	\begin{abstract}
	Our aim is to extract information about literary characters in unstructured texts.
	We employ natural language processing and reasoning on domain ontologies. 
	The first task is to identify the main characters and the parts of the story where these characters are described or act.
	We illustrate the system in a scenario in the folktale domain. 
	The system relies on a folktale ontology that we have developed based on Propp's model for folktales morphology.
	\end{abstract}
	

	\section{Introduction}
Recognising literary characters in various narrative texts is challenging both from the literary and technical perspective. 
From the literary viewpoint, the meaning of the term ``character'' leaves space to various interpretations. 
From the technical perspective, literary texts contain a lot of data about emotions, 
social life or inner life of the characters, while they are very thin on technical, straight-forward messages.
To infer the character type from literary texts might pose problems even to the human readers~\cite{bamman2014bayesian}. 
	
Interactions between literary characters contain rich social networks. 
Extracting these social networks from narrative text has gained much attention~\cite{park2013structural} in different domains such as literary fiction~\cite{elson2010extracting}, screenplays~~\cite{agarwal2014parsing}, or novels~\cite{he2013identification,agarwal2012social}.

Our aim is to correctly determine the relationships of a character in a tale and to find its role upon the development of the story. 
In line with~\cite{reiter2014nlp}, the first task is to identify the parts of the story where that character is involved.
Our approach relies on interleaving natural language processing and ontology-based reasoning.
We enact our method in the folktale domain.	
	
Information extraction systems usually have three components responsible for: named entity recognition, co-reference resolution and relationship extraction. These modules are integrated in a pipeline, in a layered manner, given that each task will use information provided by the previous neighbor.
Natural language processing has been applied in the domain of folktales~\cite{peinado2004description,fisseni2014annotating}.
	Formal models for folktales have been proposed in~\cite{lang1999declarative,propp1958morphology}.
	Character identification in folktales have been approached in~\cite{suciu2014interleaving,vannamed}.
	
	The remaining of the paper is organized as follows:
	Section~\ref{sec:ontology} presents the ontology that we developed for modeling the domain of folktales.
	Section~\ref{sec:architecture} depicts the architecture of our system.
	Section~\ref{sec:interleaving} illustrates our method to extract knowledge about characters.
	Section~\ref{sec:scenario} presents the experimental results on seven folktales. 
	Section~\ref{sec:related} browses related work, while section~\ref{sec:con} concludes the paper.
	
	\section{Engineering the folktale ontology}
	\label{sec:ontology}
	To support reasoning in the folktale domain, we developed an ontology used to extract knowledge regarding characters. 
	We assume the reader is familiarised with the syntax of Description Logic (DL). 
	For a detailed explanation about families of description logics, the reader is referred to~\cite{baader2003description}.

\begin{figure}
		\centering
		\includegraphics[scale=0.7]{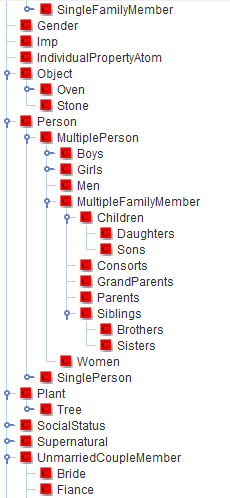}
		\caption{Folktale ontology.}
		\label{fig:ontology}
	\end{figure}

	To support character identification and reasoning on these characters we need structured domain knowledge. 
	Hence, we developed an ontology for the folktale domain as shown in Fig.~\ref{fig:ontology}. 
	Our folktale ontology formalizes knowledge from three sources:
	1) the folktale morphology as described by the Propp model~\cite{propp1958morphology};
	2) various entities specific to folktales (i.e., animals, witch, dragons); and 
	3) common family relations (i.e., child, fiancee, groom).
	In the following, these three knowledge sources are detailed:

	\paragraph{Folktale morphology}
	Firstly, we rely on the Propp's model~\cite{propp1958morphology} of the folktale domain. 
	In the Propp's model the story broke down into several sections. 
	Propp demonstrated that the sequence of sections appears in the same chronological order in Russian folktales. 
	Propp identified a set of character types that appear in most of the folktales (see Table~\ref{tab:characters}). 
	
	\begin{table}
		\caption{Main characters in the Propp's model.}
		\begin{footnotesize}
		\begin{tabularx}{\linewidth}{|p{2cm}|X| }
				\hline
				{\it Name} & {\it Description}\\
				\hline
				Villain & The opponent of the hero - often the representation of evil.\\ \hline 
				Dispatcher & The person that sends the hero into the journey, or the person that informs the hero about the villainy.\\ \hline
				(Magical) Helper & The one that helps the hero into its journey.\\ \hline
				Princess or Prize & It represents what the hero receives when it is victorious.\\ \hline 
				Donor & Prepares the hero for the battle.\\ \hline
				Hero & The main character in a story - often the representation of good.\\ \hline
				False hero & The one that tries to steal the prize from the hero, or tries to marry the princess.\\ \hline
			\end{tabularx}
		\end{footnotesize}
	\label{tab:characters}
	\end{table}

	The corresponding formalization in Description Logic appears in Fig.~\ref{fig:propp}, where the characters are divided in nine types (axiom 1).
	In axiom 2, a false hero is a hero who is also a villain.
	Axiom 3 divides the characters into negative and positive ones.
	Note that positive and negative characters are not disjoint, as for instance the concept \textit{Prisoner} belongs to both sets. 
	
	\begin{figure}
		\begin{footnotesize}
			\begin{tabular}{lp{8cm}}
				$A_1$ & Agent $\sqcup$ Donor $\sqcup$ FalseHero $\sqcup$ Hero $\sqcup$ Prisoner $\sqcup$ Villain 
				$\sqcup$ Dispatcher $\sqcup$ MagicalHelper $\sqcup$ Princess $\sqsubseteq$ Character\\
				$A_2$ & Hero $\sqcap$ Villain $\sqsubseteq$ FalseHero\\
				$A_3$ &  PositiveCharacter $\sqcup$ NegativeCharacter $\sqsubseteq$  Character\\
				$A_4$ & Villain $\sqcup$ FalseHero $\sqcup$ Prisoner  $\sqsubseteq$  NegativeCharacter\\
				$A_5$ & Hero $\sqcup$ MagicalHelper $\sqcup$ Agent $\sqcup$ Donor $\sqcup$ Prisoner $\sqcup$ Dispatcher $\sqsubseteq$  PositiveCharacter\\
			\end{tabular}
		\end{footnotesize}
		\caption{Formalising the Propp's model of folktales.}
		\label{fig:propp}
	\end{figure}

	\paragraph{Folktale main entities} Secondly, the common entities appearing in folktales were formalized
	in Fig.~\ref{fig:common}.
	The axioms depict the animals (axiom 21), witches or enchantresses which are women with a single social status (axioms 22 and 23), and supernatural characters like \textit{Giant} in axiom 24.  
	Specific characters like \textit{Goldsmith} or \textit{King}, and various objects (i.e. \textit{oven}) are also modeled. 
	A prince is defined in axiom 28 as a son that have a parent either a king or a queen. 
	Similarly, the princess is a daughter with at least on parent of type king or queen (axiom 30). 
	
	\begin{figure}
		\begin{footnotesize}
			$A_{21}$ Bear $\sqcup$ Bird $\sqcup$ Dog $\sqcup$ Duck $\sqcup$ Frog $\sqcup$ Horse $\sqcup$ Lion \ensuremath{\sqsubseteq}~SingleAnimal\\
			$A_{22}$ Enchantress~\ensuremath{\equiv}~Witch\\
			$A_{23}$ Enchantress~\ensuremath{\sqsubseteq}~Woman~$ \sqcap$~SingleSocialStatus~\\
			$A_{24}$ Giant~\ensuremath{\sqsubseteq}~Supernatural~\\
			$A_{25}$ Goldsmith~$ \sqcup$ Helmsman~ \ensuremath{\sqsubseteq}~SingleSocialStatus~\\
			$A_{26}$ King~\ensuremath{\sqsubseteq}~SingleSocialStatus~\\
			$A_{27}$ Oven~\ensuremath{\sqsubseteq}~Object~\\
			$A_{28}$ Prince~\ensuremath{\equiv}~Son~\ensuremath{\sqcap}~\ensuremath{\exists}hasParent.King~\ensuremath{\sqcup}~\ensuremath{\exists}hasParent.Queen\\
			$A_{29}$ Prince~\ensuremath{\sqsubseteq}~SingleSocialStatus~\\
			$A_{30}$ Princess~\ensuremath{\equiv}~Daughter~\ensuremath{\sqcap}~\ensuremath{\exists}hasParent.King~\ensuremath{\sqcup}~\ensuremath{\exists}hasParent.Queen\\
		\end{footnotesize}
		\caption{Common entities in the folktale domain.}
		\label{fig:common}
	\end{figure}

	\paragraph{Family relationships in folktale} Fig.~\ref{fig:family} lists part of the family relationships adapted to reason in the folktale domain. 
	A significant part of these relationships are correlated with the recurrent theme of the main character who is finding his bride or fiancee. 
	
	\begin{figure}
		\begin{footnotesize}
			$A_{50}$ Boy~\ensuremath{\sqsubseteq}~SinglePerson~\\
			$A_{51}$ Boys~\ensuremath{\sqsubseteq}~MultiplePerson~\\
			$A_{52}$ Bride~\ensuremath{\equiv}~Fiancee\\
			$A_{53}$ Bride~\ensuremath{\sqsubseteq}~UnmarriedCoupleMember~\\
			$A_{54}$ Brother~\ensuremath{\sqsubseteq}~Sibling~$\sqcap$ Male\\
			$A_{55}$ Daughter~\ensuremath{\equiv}~Girl~\ensuremath{\sqcap}~\ensuremath{\sqcap}Child~\ensuremath{\exists}hasParent.Parent\\
			$A_{56}$ Father~\ensuremath{\equiv}~Man~\ensuremath{\sqcap}~\ensuremath{\exists}hasChild.Child\\
			$A_{57}$ Fiance~\ensuremath{\equiv}~Groom\\
			$A_{58}$ Fiance~\ensuremath{\sqsubseteq}~UnmarriedCoupleMember~\\
			$A_{59}$ Fiancee~\ensuremath{\equiv}~Bride\\
			$A_{60}$ Fiancee~\ensuremath{\sqsubseteq}~UnmarriedCoupleMember~\\
			$A_{61}$ Girl~\ensuremath{\equiv}~Maiden\\
			$A_{62}$ Girl~\ensuremath{\sqsubseteq}~SinglePerson~\\
			$A_{63}$ Husband~\ensuremath{\sqsubseteq}~Consort~\\
		\end{footnotesize}
		\caption{Family relationships in the folktale domain.}
		\label{fig:family}
	\end{figure}
	
	To facilitate reasoning on the ontology, we allow several extensions of the $\mathcal{ALC}$ version of description logics~\cite{baader2003description}.
		Using role inheritance we can specify that the role \textit{hasFather} is more specific than the role \textit{hasParent}. 
	Hence, if we find in the folktale that a character has a father, the system deduces based on role inheritance that the character has also a parent. 
	Similarly, inverse roles like \textit{hasChild} and \textit{hasParent} are used to infer new knowledge based on the partial knowledge extracted by natural language processing. 
	If we identify that two individuals are related by the role \textit{hasChild}, the system deduces that those individuals are also related by the role \textit{hasParent}. 
	The domain restriction specifies that only persons can have brothers. 
	The range restriction constraints the range of the role \textit{hasGender} to the concept \textit{Gender}. 
	
	\begin{table}
		\caption{Exploiting role constraints to reason on the ontology.}
		\begin{footnotesize}
			\begin{tabular}{|p{2.2cm}|l|}\hline
				\textit{Extensios of ALC}& \textit{Folktale examples}\\ \hline
				Role inheritance & hasBrother~\ensuremath{\sqsubseteq}~hasSibling, hasFather~\ensuremath{\sqsubseteq}~hasParent,\\ 
				& hasHusband~\ensuremath{\sqsubseteq}~hasConsort\\ \hline
				Inverse roles&  hasHusband~\ensuremath{\equiv}~hasWife\ensuremath{^-}, hasChild~\ensuremath{\equiv}~hasParent\ensuremath{^-}\\ \hline
				Transitive roles & hasSibling\ensuremath{^t}\\ \hline
				Domain restriction & \ensuremath{\exists}hasBrother.$\top$~\ensuremath{\sqsubseteq}~Person\\ \hline
				Range restriction & \ensuremath{\top}~\ensuremath{\sqsubseteq}~\ensuremath{\forall}hasBrother.Person, \ensuremath{\top}~\ensuremath{\sqsubseteq}~\ensuremath{\forall}hasGender.Gender\\ \hline
				symmetric roles & hasConsort~\ensuremath{\equiv}~hasConsort\ensuremath{^-}\\ \hline
				cardinality constraints & \ensuremath{\top}~\ensuremath{\sqsubseteq}~\ensuremath{\leq}1~hasGender.Thing\\ \hline
			\end{tabular}
		\end{footnotesize}
		\label{tab:extensions}
	\end{table}

	\section{System architecture}
	\label{sec:architecture}
	
	\begin{figure*}
		\centering
		\includegraphics[scale=0.77]{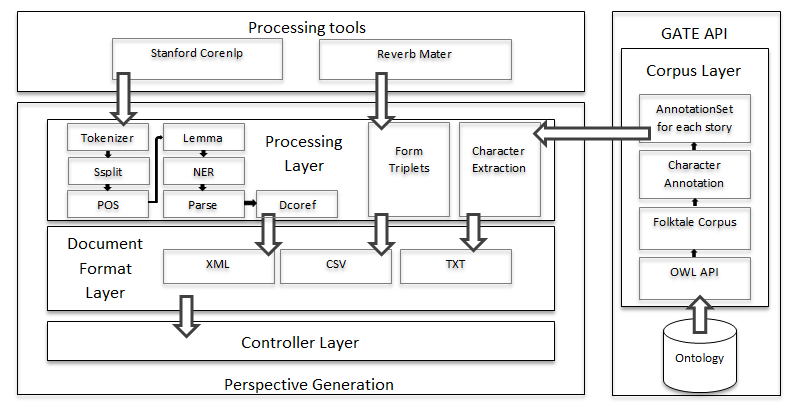}
		\caption{The System Architecture}
		\label{fig:architecture}
	\end{figure*}
	
Extracting knowledge about characters is obtained by interleaving natural language processing (NLP) and reasoning on ontologies. 
The NLP component is based on GATE text engineering tool~\cite{bontcheva2004evolving}, while reasoning in DL on the OWLAPI~\cite{horridge2011owl}, as depicted by the architecture in Fig.~\ref{fig:architecture}.
	
Firstly, the folktale ontology is processed using OWLAPI to generate classes of characters from the ontology into GATE. 
	The folktale corpus is analysed aiming to populate the ontology and to annotate each folktale with the identified named entities. 
In parallel to the annotation process, the Stanford parser creates the coreference information files. 
	The task is challenging, as even a human might have a problem in decoreferencing some of the sentences, as  example~\ref{ex:smiths} illustrates.
	
	\begin{example}
		"The Smiths went to visit the Robertsons. 
		After that, they stayed home, watching tv.", where \textit{"they"} might be tied to the Smiths, or the Robertsons, or to both of the families. 
		\label{ex:smiths}
	\end{example}
	
	For de-coreferencing, the following pipeline was designed (left part of Fig.~\ref{fig:architecture}).
	The tokenizer groups all the letters into words. 
	Next, the sentence splitter (Ssplit) groups the sequence of tokens obtained in the previous step into sentences. 
	The part of speech (POS) annotation labels all the tokens from a sentence with their POS tags. 
	Lemma annotation generates the word lemmas for all the tokens in the corpus. The next step is to apply named entity recognition (NER) so that the numerical and temporal entities are recognized. 
	This is done using a conditional random fields (CRF) sequence taggers trained on various corpora. 
	The parse function provides a full syntactic analysis for each sentence in the corpora. 
	Finally, the coreference chain annotation (Dcoref) obtains  both the pronominal and nominal coreference resolution.
	After coreference resolution, the stories are updated with the coreference information. 
	
	\begin{table*}
		\caption{Extracting triplets from folktales using Reverb.}
		\begin{footnotesize}
			\begin{tabularx}{\hsize}{|X|p{1cm}|p{1cm}|p{1cm}|p{1cm}|X|X|} \hline
			 {\it Original Sentence} & {\it Nominal Phrase (arg1)} & {\it Verb Phrase (arg2)} & {\it Nominal Phrase (arg3)} & {\it Extraction Confidence} & {\it POS tags} & {\it Chunk tags}\\ \hline
			 {\it 10} & {\it 3} & {\it 4} & {\it 5} & {\it 9} & {\it 11} & {\it 12}\\ \hline
			 Good heavens, said the girl, no strawberries grow in winter. & no strawberries & grow in & winter & 0.505 & JJ NNS , VBD DT NN , DT NNS VB IN NN . & B-NP I-NP O B-VP B-NP I-NP O B-NP I-NP \\ \hline
			 The king's daughter began to cry , for daughter was afraid of the cold frog which daughter did not like to touch, and which was now to sleep in daughter pretty, clean little bed. & daughter & was afraid of & the cold frog & 0.691 & DT NN POS NN VBD TO VB , IN NN VBD JJ IN DT JJ NN WDT NN VBD RB IN TO VB , CC WDT VBD RB TO VB RP NN RB , JJ JJ NN . & B-NP I-NP I-NP I-NP B-VP I-VP I-VP O B-PP B-NP B-VP B-ADJP B-PP B-NP I-NP I-NP B-NP I-NP B-VP O O B-VP I-VP O O B-NP B-VP B-ADVP B-VP I-VP B-NP I-NP B-ADVP O B-NP I-NP I-NP O\\ \hline
			 When everything was stowed on board a ship, faithful John put on the dress of a merchant, and the king was forced to do the same in order to make king quite unrecognizable. & John & put on & the dress of a merchant & 0.876 & WRB NN VBD VBN IN NN DT NN , NN NNP VBD IN DT NN IN DT NN , CC DT NN VBD VBN TO VB DT JJ IN NN TO VB NN RB JJ . & B-ADVP B-NP B-VP I-VP B-PP B-NP B-NP I-NP O B-NP B-NP B-VP B-PP B-NP I-NP I-NP I-NP I-NP O O B-NP I-NP B-VP I-VP I-VP I-VP B-NP I-NP B-SBAR O B-VP I-VP B-NP B-ADJP I-ADJP O \\ \hline
			 Sons each kept watch in turn, and sat on the highest oak and looked towards the tower. & each & kept watch in & turn & 0.880 & NNPS DT VBD NN IN NN , CC VBD IN DT JJS NN CC VBD IN DT NN . & O B-NP B-VP B-NP B-PP B-NP O O B-VP B-PP B-NP I-NP I-NP O B-VP B-PP B-NP I-NP O \\ \hline
			 Rapunzel grew into the most beautiful child under the sun. & Rapunzel & grew into & the most beautiful child & 0.830 & NNP VBD IN DT RBS JJ NN IN DT NN . &  B-NP B-VP B-PP B-NP I-NP I-NP I-NP B-PP B-NP I-NP O \\ \hline
			  The king's son ascended, but instead of finding son dearest rapunzel, son found the enchantress, who gazed at son with wicked and venomous looks. & the enchantress & gazed at & son & 0.586 & DT NN POS NN VBD , CC RB IN VBG NN NN NN , NN VBD DT NN , WP VBD IN NN IN JJ CC JJ NNS . & B-NP I-NP I-NP I-NP B-VP O O B-PP I-PP B-VP B-NP I-NP I-NP O B-NP B-VP B-NP I-NP O B-NP B-VP B-PP B-NP B-PP B-NP I-NP I-NP I-NP O \\ \hline
			\end{tabularx}
		\end{footnotesize}
		\label{tab:reverbOutput}
	\end{table*}
	
	The Reverb information extraction tool~\cite{fader2011identifying} is used to generate triplets containing the following structure: $\langle$nominal phrase, verb phrase, nominal phrase$\rangle$.
	For the sentence \textit{"Good heavens, said the girl, no strawberries grow in winter"}, the output of Reverb is exemplified in Table~\ref{tab:reverbOutput}.  
		In order to obtain the triplets, each sentence has to be POS-tagged and NP-chunked.

	\section{Interleaving natural language processing with reasoning on ontologies}
	\label{sec:interleaving}
	
	\begin{figure}
		\centering
		\includegraphics[scale=0.40]{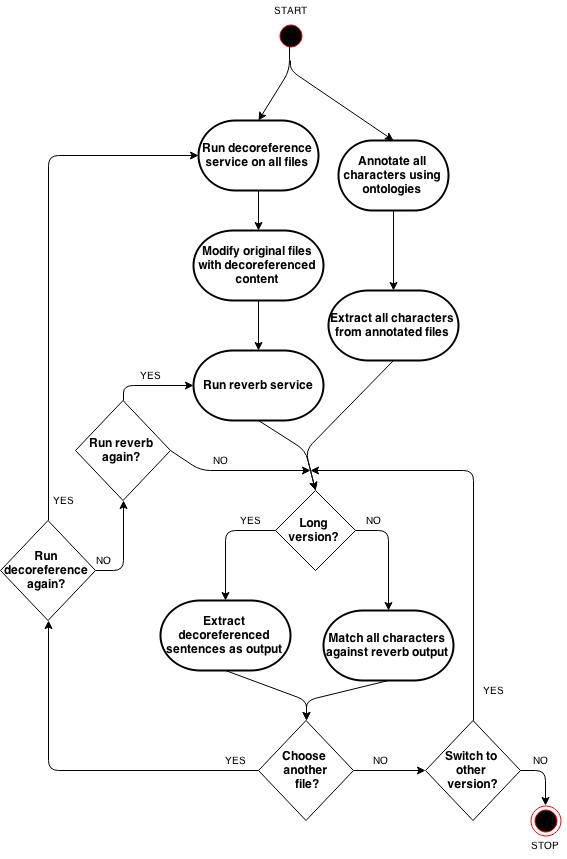}
		\caption{Main execution phases.}
		\label{fig:Flowchart}
	\end{figure}

	This section details three algorithms used to identify knowledge about characters. 
	Algorithm~\ref{alg:processing2} identifies characters in the folktale.
	Algorithm~\ref{alg:processing} is used for anaphora resolution of the named entities recognized as characters.
	Algorithm~\ref{alg:processing1} extracts knowledge about characters from the de-coreferences. 
	The execution flow of this pipeline, is presented in Fig.~\ref{fig:Flowchart}.

	\begin{algorithm}
		\SetKwInOut{Input}{Input}
		\Input{
			$O_f$ - Folktale ontology;\newline
			$S$ - Corpus of folktales;\newline
			$JN$ - Jape rules to identify definite and indefinite nominal phrases;\newline
			$JC$ - Jape rules to identify candidate characters;\newline
			$JR$ - Jape rules to identify character's relation to the ontology;}
		\KwResult{$C$: Set of annotated characters;}
		\BlankLine
		$C\leftarrow \emptyset$\;
		$NP\leftarrow$ applyRules($JN$, $S$)\;
		\While{applyRules($JC$, $S$, $NP$) $\neq$ null}{
			$NC \leftarrow$ applyRules($JC$, $S$, $NP$)\;
			$Rel\leftarrow$ applyRules($JR$, $S$, $NC$)\;
			\ForEach{$r\in Rel$}{
				\ForEach{$concept$ from $r$}{
					\If{checkCast($NC$, $concept$)}{
						cast($NC$, $concept$)\;}
				}}
				\While{is\_referred($S$, $NC$)}{
					$Ref$ = getReference()\;
					link($NC$, $Ref$)\;}
				$C \leftarrow C \cup NC$\;
			}
			\caption{Character extraction algorithm.}
			\label{alg:processing2}
		\end{algorithm}
		
		Natural language processing is enacted to populate the folktale ontology.
		The extraction Algorithm~\ref{alg:processing2} is performed repetitively on a document, each time using the newly populated ontology file. 
		In this way, the algorithm interleaves reasoning on ontology with natural language processing based on Japes rules~\cite{thakker2009gate}.
		The first step is to apply the Jape rules $JN$ on the folktale corpus aiming to identify all the definite and indefinite nominal phrases. 
		Given that the characters are nominal phrases, this first step returns all the information needed, plus some extra phrases that have to be filtered out. 
		
		Next, the Jape rules $JC$ are enacted to select candidate characters from the set of nominal phrases previously identified.
		For each character found, a set of rules $JR$ is used to match the character against a concept in the ontology.
		
		After identifying a concept for which the character is an instance, the algorithm exploits reasoning on ontology to identify all atomic concepts to which the character belongs. 
		For instance, a character identified as $Daughter$ will be an instance of $Girl$, $Child$, $Maiden$, $SinglePerson$ (recall Fig.~\ref{fig:family}).
		For each concept to which the character belongs, the algorithm looks again in the corpus to see if there are other mentions of the newly introduced character. 
		If this is the case, the character is related with the new knowledge.

		\begin{algorithm}
			\SetKwInOut{Input}{Input}
			\Input{
				$S$: Corpus of folktales;
				$P$: Pipeline configuration for decoreferencing;\newline
				$FN$: List with filenames for each $S$;\newline
				$SC$: Stanford-CoreNLP command;}
			\KwResult{$D$: Decoreferenced texts of files from $FN$;}
			\BlankLine
			$Files$ = run($SC$, $P$, $FN$)\;
			\ForEach{$file$ in $Files$}{
				$D\leftarrow S$\;
				\ForEach{$coref\_group \in file$}{
					$rep\leftarrow$ findRepresentative($coref\_group)$\;
					\ForEach{$coref\_word \in coref\_group$}{
						replace($D$, $coref\_word$, $rep$ )\;}
				}
			}
			\caption{Decoreference algorithm.}
			\label{alg:processing1}
		\end{algorithm}

		\begin{algorithm}
			\SetKwInOut{Input}{Input}
			\Input{$R$: Reverb command;\newline
				$V$: The version indicator. True if long version, false otherwise;\newline
				$C$: Set of characters resulted from algorithm~\ref{alg:processing2};\newline
				$D$: Decoreferenced text resulted from algorithm~\ref{alg:processing1};}
			\KwResult{$P$: String containing character's perspective in $S$;}
			\BlankLine
			$RR$ = run($R$, $D$)\;
			\eIf{$V$ = true}{
				\ForEach{$c \in C$}{
					\ForEach{$line \in RR$}{
						$sentence\leftarrow$ getSentence($line$)\;
						\If{$c \in sentence$}{
							$P \leftarrow P \cup sentence$\;}
					}
				}
			}{
			\ForEach{$c \in C$}{
				\ForEach{$line \in RR$}{
					$triplet\leftarrow$ getTriplet($line$)\;
					\If{$c \in triplet$}{
						$P \leftarrow P \cup triplet$\;}
				}
			}
		}
		\caption{Finding character's perspective.}
		\label{alg:processing}
	\end{algorithm}

	The decoreferencing algorithm (Alg.~\ref{alg:processing1}) uses as input the processing pipeline and the folktale corpus. 
	The basic processing steps needed are the following: \textit{tokenize, ssplit, pos, lemma, ner, parse, dcoref}. 
	The decoreferencing algorithm is run on all stories at once, but it generates different output file for each story represented by the filename. 
	In the first step, the Stanford parser applies the execution pipeline on the corpora of folktales.
	For each resulted file, the algorithm searches for coreference groups. 
	In order to be able to return the modified text, the original text has to be stored in the returning argument of the algorithm. 
	For each coreference group found, firstly the referenced word has to be processed and kept into a variable
	and then, each coreferenced word found, belonging to the group, has to be replaced in the original text with the referenced variable. 
	In the end, the decoreferenced text for each corpus file is obtained.
	
	Algorithm~\ref{alg:processing} takes as input the result of algorithms~\ref{alg:processing2} and (alg~\ref{alg:processing1}. 
	The set of characters is used as the input, while the decoreferenced texts are used as an environment from which the algorithm extracts the perspective. 
	For each character in the set of characters resulted from the extraction algorithm (alg~\ref{alg:processing2}), each line that resulted from reverb execution is processed. 
	From each line, the sentence is extracted based on the output format of the Reverb service presented in Table~\ref{tab:reverbOutput}. 
	If the character, from the character set, is mentioned in the sentence, then the sentence is appended to the output variable. 
	These columns are combined in a triplet, and  it is checked to see whether the current character appears is present in this triplet. 
	In this case, the triplet is appended to the output variable. 
	This algorithm’s score is represented by a subunitary number that represents the confidence that the extraction was correct. 
	
	\section{Experimental results}
		\label{sec:scenario}
	
	\subsection{Running scenario}
	The system was tested against seven stories (Table~\ref{tab:test}). 
	This section illustrates the results of this pipeline for the secondary character Henry from the story ``The frog king''.  
	The fragment on which the algorithms were applied is listed in Example~\ref{ex:frog}.
	
	\begin{example}
		"Then they went to sleep, and next morning when the sun awoke them, a
		carriage came driving up with eight white horses, which had white
		ostrich feathers on their heads, and were harnessed with golden
		chains, and behind stood the young king's servant Faithful Henry.
		Faithful Henry had been so unhappy when his master was changed into a
		frog, that he had caused three iron bands to be laid round his heart,
		lest it should burst with grief and sadness.  The carriage was to
		conduct the young king into his kingdom.  Faithful Henry helped them
		both in, and placed himself behind again, and was full of joy because
		of this deliverance.  And when they had driven a part of the way the
		king's son heard a cracking behind him as if something had broken.
		So he turned round and cried, "Henry, the carriage is breaking."
		"No, master, it is not the carriage.  It is a band from my heart,
		which was put there in my great pain when you were a frog and
		imprisoned in the well."  Again and once again while they were on
		their way something cracked, and each time the king's son thought the
		carriage was breaking, but it was only the bands which were springing
		from the heart of Faithful Henry because his master was set free and
		was happy."
		\label{ex:frog}
	\end{example}

	\begin{table}
		\caption{Eliciting knowledge about Henry.}
		\begin{center}
			\begin{tabular}{|l|l|}\hline
				& \textit{Character: Henry}\\ \hline
				1& Henry master was changed into a frog\\ \hline
				2& Henry had caused three iron bands\\ \hline
				3& faithful Henry helped bands\\ \hline
				4& bands placed Henry\\ \hline
				5& Henry was full of joy\\ \hline
				6& the bands were springing from the heart of faithful Henry\\ \hline 
			\end{tabular}
		\end{center}
		\label{tab:henry}
	\end{table}
	
	The method has two kind of results - one for the long version, and one for the short version. Firstly, the results for the short version are listed in Table~\ref{tab:henry}.
	Note that the output text is the decoreferenced one - this is the reason why the character might talk about itself in third person. 
	Because of the de-coreferenced version of the stories part of text might not be correct from the human reader perspective. 
	But it is the easiest way to understand the context of a character. 
	Otherwise, it would be hard to see that when the text says "his master", that "his" refers to Henry, as Example~\ref{ex:henry} bears out. 
	
	\begin{example}
		1. Then companion went to sleep, and next morning when the sun awoke companion, a band came driving up with eight white horses, which had white ostrich feathers on companion heads, and were harnessed with golden chains, and behind stood the young king's servant faithful Henry.
		
		2. Faithful Henry had been so unhappy when henry master was changed into a frog, that Henry had caused three iron bands to be laid round henry heart, lest heart should burst with grief and sadness.
		
		3. Faithful Henry helped bands both in, and placed Henry behind again, and was full of joy because of this deliverance.
		
		4. Again and once again while you were on you way something cracked, and each time the king's son thought the band was breaking , but it was only the bands which were springing from the heart of faithful Henry because Henry master was set free and was happy."
		
		\label{ex:henry}
	\end{example}

	There are some cases in which there will be no result for a character (Example~\ref{ex:modified}). 
	Given that the character was extracted from the original file, by using Algorithm~\ref{alg:processing2}, there is a certainty that the character exists in the story.
	
	\begin{example}
		When trying to search for the perspective of character \textit{"waiting-maid"} in the story \textit{"Faithful John"}, the application will not be able to find any solution. In the unmodified text, the \textit{son} character is introduced in the following way: \textit{"She took him by the hand and led him upstairs, for she was the waiting-maid."}
		\label{ex:modified}
	\end{example}

	This happens because, when the anaphoric decoreference is run (Algorithm~\ref{alg:processing1}), the file is changed in the following way: \textit{"Girl took oh by the hand and led oh upstairs , for girl was the girl ."}. The change happened because the decoreferencing tool interpreted \textit{"the waiting-maid"} as being tied up to the word \textit{"she"}, and, which is tied to \textit{"the girl"} from the following phrase \textit{"Then said the girl ` the princess must see these , girl has such great pleasure in golden things , that girl will buy all you have . '"}. In this way, this character's part will be attributed to the \textit{"girl"}, which is the main character of the story. 
	This situation in which the story is talking about a general character, but only after the main events, the character is finally revealed, is called \textit{cataphora}~\cite{kazanina2010differential}.
	

	\begin{table}
		\caption{Accuracy of the algorithms.}
		\begin{center}
			\begin{tabular}{|l|l|}\hline
				{\it Story} & {\it Accuracy}\\
				\hline
				The Magic Swan-Geese& 75\%\\ \hline
				The Frog King& 62\%\\ \hline
				The King's Son who Feared Nothing& 76\%\\ \hline
				Faithful John& 63\%\\ \hline
				The Twelve Brothers& 65\%\\ \hline
				Rapunzel& 74\%\\ \hline
				The Three Little Men in the Woods& 73\%\\ \hline 
				{\it Average} & 70\% \\ \hline 
			\end{tabular}
		\end{center}
		\label{tab:test}
	\end{table}

	\subsection{Accuracy of the method}
	
	The accuracy of our method is influenced by:
	1) accuracy of character identification; 
	2) accuracy of identifying co-references; 
	3) accuracy of Reverb when extracting triplets (the confidence indicator). 
	Each of this services has an accuracy error that will be propagated from one component to another. 
	We performed various tests on the corpus used for character identification, and we obtained an average accuracy of 70\% (Table~\ref{tab:test}). 
	When calculating the accuracy, 20 characters were taken into consideration, meaning that for each story, about 3 characters were chosen. 
	These characters were manually selected from the set of characters output by the character extraction system presented in~\cite{suciu2014interleaving,vannamed}. 
	The characters were selected by choosing 2 main characters and a secondary character for each story. 

\begin{figure}
		\centering
		\includegraphics[scale=0.6]{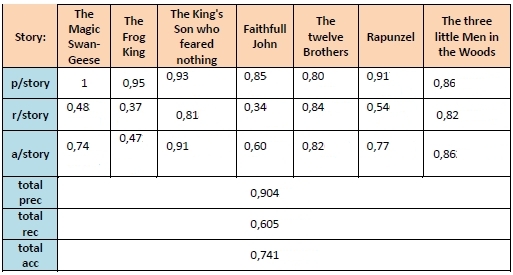}
		\caption{Precision, recall, and accuracy for the seven folktales analyzed.}
		\label{fig:results}
	\end{figure}

\begin{figure}
		\centering
		\includegraphics[scale=0.65]{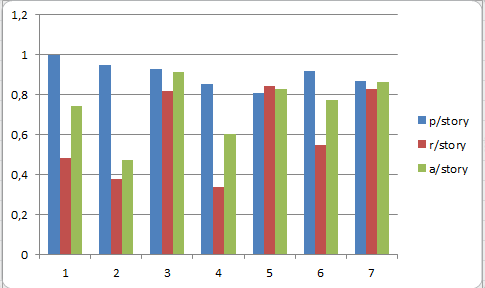}
		\caption{Comparing precision, recall and accuracy for each story.}
		\label{fig:results2}
	\end{figure}

	The testing was performed on seven different stories, and for each story, a set of main characters was chosen. The obtained overall accuracy is 74\%, having an overall precision of 90\% and a recall of 60\%. 
	The results are presented in Fig.~\ref{fig:results}.
	Figure~\ref{fig:results2} depicts the distribution of precision, recall and accuracy over the stories. The values were calculated using the following formulas: 
	
	\begin{center}
		$precision$ = $tp \over tp + fp$
		
		$recall$ = $tp \over tp + fn$
		
		$accuracy$ = $tp + tn \over tp + fp + tn + fn$
	\end{center}
	
	where $tp$ means \textit{true positive}, and represents the number of sentences that are found both in the manually annotated set and the test set, $tn$ means \textit{true negative} and represents the number of sentences that are neither in the manually annotated set, nor in the test set, $fp$ means \textit{false positive} and represents the number of sentences that are in the test set and not in the manually annotated set, and $fn$ means \textit{false negative} and represents the number of sentences that are in the manually annotated set, but not in the test set. In the folktale context, the $tp$ represents the number of sentences that belong to the character's perspective, all those sentences that involve the character in any way.
	
	The average F-score for the Stanford-CoreNLP of 59.5 influences greatly the performance of the algorithm, as the character’s perspective cannot be extracted, given that the character is not seen as being part of the sentence.
	The accuracy can be improved if a better decoreferencing tool will be used. Other coreference tools are 
	For the anaphoric decoreference, there are several other tools (BART, JAVARAP, GuiTar and ARKref), but, from all, the Stanford-CoreNLP has be highest accuracy percentage.

	There is ongoing research in the coreference resolution domain, 
	When calculating the performance scores, the extraction of the correct sentence was considered, and not on the correctness of the extracted sentence. 
	Even though the right sentence was extracted, the information in the sentence will be according to the coreference resolution result. 
	Hence, an error might be observed when reviewing the structure of the sentences.
	The algorithm’s performance is also influenced by the scores obtained by the Reverb tool. 
	Also, the named entity recognition has an average precision of 79\% and a recall of 72\%. 
	These scores do not influence directly the algorithms performance, but they have an effect on the number of characters for which the algorithm will try to find the roles they have on the development of the story.
	Together, all these scores combined, give the performance scores of the characters perspective in texts.
	
	The current version does not extract information about the characters' roles. 
	The information extracted consists of the character identification, that is presented in ~\cite{suciu2014interleaving,vannamed}, and the story involving the character. 
	The story can be presented in a standardized version.

	\section{Discussion}
	\label{sec:related}

	We can enact our solution in other domains instead of folktales. We exemplify he following three domains: 
	a) software requirements, b) marketing and c) medical domain.
	
	Consider the domain of \textit{software requirements}, where these requirements are written in natural language. 
	Our system will support the identification of various actors appearing in the requirements document. 
	First, one needs to replace the folktale ontology with a requirement ontology that provides knowledge on use cases, actors, their roles, etc. 
	The same pipeline will be used to: 
	1) identify main actors (admin, various users, etc) and
	2) extract knowledge about various actions these actors are supposed to perform. 
	
	Another domain that could benefit from the same pipeline of execution, would be the \textit{marketing domain}. 
	Consider a dataset of product reviews or accommodation places in the tourism domain~\cite{varga2011integrating}.
	The system would extract only the sentences that reference the mentioned item. 
	By having access to all the sentences of interest, further analysis is facilitated without having to process the entire text.

	Similar extraction systems have been proposed for the \textit{medical domain} to extract information from clinical narratives.
	In this line, the MedEx system~\cite{xu2010medex} aims to extract the medication information from clinical narratives. 
	Similarly, there is also the OpenClinical system for assisting health care providers.
	
	In our approach, the extraction algorithm part is separated from the perspective searching part. 
	Therefore, any ontology and any document can be used in order to find the character's or object's perspective in the document. 
	
	We tested our method only on seven stories. 
 With a complexity of $O(n^3)$ in sentence length of syntactic parsing, our syntactic based on Stanford parser might be too slow for large corpus as the one of 15099 narratives analysed in~\cite{bamman2014bayesian}.

	\section{Conclusions}
	\label{sec:con}
	
	Our method is able to extract knowledge on various characters. 
	Our current accuracy for information extraction in the folktale domain is 74\%. 
	The experimental results were obtained for seven stories in the folktale domain.
	The precision score is above 90\%, 
	With an overall recall of only 60\%, there are high chances that 
	not all the information regarding a product was extracted.
	
	The developed algorithms aggregate three different services: 
	Firstly, the named entity recognition was implemented by using an ontology based on Propp’s formal model. 
	Based of this ontology, and some implemented Jape rules, 
	the characters are extracted from a given story. 
	Secondly, a coreference resolution tool was implemented by enacting anaphoric resolution to 
	eliminate co-referenced words and to replace them with their representative,
	Thirdly, finding relationships between characters was integrated in order to link two noun phrases with a verbal phrase.

	\section*{Acknowledgments}
	We thank the reviewers for their valuable comments.
	Part of this work was supported by the Department of Computer Science of Technical University of Cluj-Napoca, Romania.
	
	\bibliographystyle{IEEEtran}
	\bibliography{cop} 
	
\end{document}